%% file: paper.tex
\documentclass[letterpaper, 10 pt, conference]{ieeeconf}  

\IEEEoverridecommandlockouts                              

\usepackage{graphicx}
\usepackage{subcaption}
\usepackage{amsmath} 
\usepackage{amssymb}  
\usepackage{url}
\usepackage{xcolor}

\title{\LARGE \bf
An Approach to Vehicle Trajectory Prediction Using Automatically Generated Traffic Maps
}

\author{Jannik Quehl$^1$, Haohao Hu$^1$, Sascha Wirges$^2$ and Martin Lauer$^1$
\thanks{*The research leading to these results has received funding from the German collaborative research center ``SPP 1835 - Cooperative Interacting Automobiles'' (CoInCar) granted by the German Research Foundation (DFG).}
\thanks{$^1$Authors are with Institute of Measurement and Control Systems,
        Karlsruhe Institute of Technology, Karlsruhe, Germany
        {\tt\small \{quehl,haohao.hu,martin.lauer\}@kit.edu}}%
\thanks{$^2$Author is with FZI Research Center for Information Technology, Karlsruhe, Germany.
        {\tt\small wirges@fzi.de}}%
}

\begin{document}
	\onecolumn
	\LARGE{IEEE Copyright Notice}
	
	Copyright (c) 2018 IEEE
	Personal use of this material is permitted. Permission from IEEE must be obtained for all other uses, in
	any current or future media, including reprinting/republishing this material for advertising or promotional
	purposes, creating new collective works, for resale or redistribution to servers or lists, or reuse of any
	copyrighted component of this work in other works.\\
	
	Accepted to be Published in: Proceedings of the 2018 IEEE Intelligent Vehicles Symposium (IV'18), June 26 -- June 29, 2018
	\normalsize
	\twocolumn

\newpage

\maketitle
\thispagestyle{empty}
\pagestyle{empty}

\input{tex/abstract}
\input{tex/introduction}
\input{tex/graph_creation}
\input{tex/prediction}
\input{tex/evaluation}
\input{tex/conclusions}

\bibliographystyle{IEEEtran}
\bibliography{IEEEabrv,tex/references}

\end{document}

%% file: tex/abstract.tex
\begin{abstract}
Trajectory and intention prediction of traffic participants is an important task in automated driving and crucial for safe interaction with the environment.
In this paper, we present a new approach to vehicle trajectory prediction based on automatically generated maps containing statistical information about the behavior of traffic participants in a given area. 
These maps are generated based on trajectory observations using image processing and map matching techniques and contain all typical vehicle movements and probabilities in the considered area. Our prediction approach matches an observed trajectory to a behavior contained in the map and uses this information to generate a prediction.
We evaluated our approach on a dataset containing over 14000 trajectories and found that it produces significantly more precise mid-term predictions compared to motion model-based prediction approaches.
\end{abstract}

%% file: tex/introduction.tex
\section{Motivation and Related Work}

In automated driving, planning and understanding the ego trajectory is one of the most important and fundamental tasks.
It is necessary in order to plan emergency maneuvers, enable automatic lane keeping, dynamically adapt the velocity or in order to drive autonomously.
To solve this task it is necessary to know how the car's surroundings will change during the planning horizon.
Since these changes cannot be measured or known in advance it is necessary to perform some form of prediction.
For static surroundings or obstacles this task is trivial.
However, for moving objects and other traffic participants in particular the estimation of future behavior becomes a very challenging problem.
In commonly used motion planning tasks, the behavior of traffic participants is fully expressed by their trajectories.
Therefore, we focus on trajectory prediction in this work.

Whereas pedestrians and in many cases also cyclists move arbitrarily on free areas, roadways restrict vehicle motion to certain trajectory patterns.
Thus, we target the identification and extraction of vehicle trajectory patterns in this work and present a framework to predict vehicle trajectories.

Vehicle trajectory prediction is not an exact and deterministic problem since an observer does usually not have all the relevant information such as a driver's intention or driving style.
Many prediction approaches make the assumption that these factors cannot be reliably estimated and therefore only use previously known information about vehicle dynamics and the current movement state of the observed vehicle for their trajectory prediction.
For example, it is known that given a current velocity the other vehicle cannot accelerate or decelerate faster than the engine or breaks physically allow and that further the driver aims too keep some degree of comfort for all passengers.
Therefore the future vehicle velocity and acceleration can be estimated within certain bounds.
A similar assumption can be made about the curvature the vehicle can drive based on the maximum steering angle and vehicle stability.
However, these bounds still do not limit the amount of trajectories to a size for which an accurate prediction is possible.

Previous work by Schubert et al. \cite{schubert2008comparison} showed that assuming a Constant Yaw Rate and Acceleration (CYRA) model provides good results for vehicle tracking tasks.
This suggests that for short-term prediction, e.g. the time between two vehicle detections, such a simple model based on vehicle dynamics provides good predictions.
This was applied for example in \cite{tamke2011flexible} and \cite{berthelot2011handling} for short-term vehicle trajectory prediction.
For long-term predictions on the other hand such predictions can become quite inaccurate since they cannot predict changes in the yaw rate that each person would be able to predict e.g. when leaving or entering a bend.
A different approach was pursued in \cite{sung2012trajectory} and \cite{laugier2011probabilistic} which incorporated a Maneuver Recognition Module (MRM) in order to identify actions from a certain set of maneuvers and using that for trajectory prediction.

\begin{figure}[t]
	\centering
	\includegraphics[width = 0.90\columnwidth]{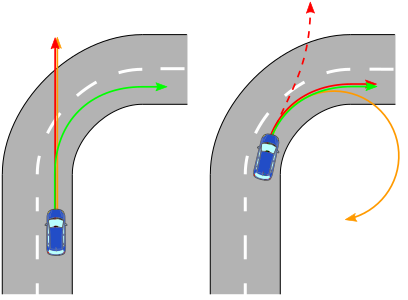}
	\caption{Comparison of different prediction approaches at two points in time.
		CYRA in orange, MRM-based prediction in red (solid and dashed) and our proposed method in green.}
	\label{fig:pred_comp}
\end{figure}

Houenou et al. \cite{houenou2013vehicle} combined motion model and maneuver based predictions trying to benefit of both approaches.
A disadvantage with maneuver based approaches, however, is that the maneuvers recognized by the MRM may be predictable a lot sooner.
A MRM for example needs a short time in order to recognize that a car starts entering a bend even though any person that sees that a bend is coming could make a better prediction far sooner. 

This paper proposes a trajectory prediction method that uses a special graph based map representation generated through trajectory observations to predict future changes in maneuver and trajectory.
By using information contained in a map this approach is able to predict driving maneuvers along the pathway of lanes and intersections and assign probabilities to different driving behavior.
Our prediction method is facilitated by statistical information about natural driving behavior in the surrounding area. 
Figure~\ref{fig:pred_comp} illustrates the different approaches with a simple example: Before the car starts turning neither an MRM nor a motion model based approach can make a correct prediction.
In the next timestep an MRM-based method might yield a correct prediction, however if a wrong maneuver is estimated (lane change instead of turning), it might still make a wrong prediction.
CYRA on the other hand will provide a good short-term but a bad long-term prediction either way.

First, we provide an overview on our automatic map creation in Section~\ref{sec:Graph Creation}.
Based on this map information, we present our novel prediction method in Section~\ref{sec:Prediction}.
By evaluating our method in Section~\ref{sec:Evaluation}, we show that it is capable to make accurate mid-term predictions.
Finally, we conclude our work in Section~\ref{sec:Conclusions}.

%% file: tex/graph_creation.tex
\section{Graph Creation}
\label{sec:Graph Creation}

When predicting vehicle trajectories it is reasonable to assume that all observed vehicles will behave according to traffic rules.
Their movement is confined to the boundaries of the road and their trajectories can be segmented into shorter trajectories with a well defined starting and endpoint.

For example consider a group of vehicles driving from on the same lane towards an intersection: A part of these vehicles will go straight while some vehicles will take a turn.
The accumulated trajectories of these different driving behaviors form a traffic network that describes the geometrical structure and drivable areas of the considered intersection.
This traffic network can be described in the form of a directed topological graph $G = (E,V)$ with transition probabilities.
These transition probabilities represent the general tendencies to choose the respective edge as basis for the next partial trajectory and can depend on a number of factors like velocity or even the current time of day (rush hour). 

In order to create such a graph, the first step is to determine the main structure of $G$, which consists of a set of nodes $V$ (or vertices) and a set of edges $E$.
This can be realized in several ways: The first method would be to use existing maps readily available online to extract information about the surrounding road structure.
Open Street Map \cite{OpenStreetMap} for example allows the automatic retrieval of such information that can be parsed to a suitable graph.
However this method has the drawback that maps like these are usually not geo-referenced precise enough everywhere to constitute a lane accurate depiction of the course of the road.
Further they are often missing relevant information like the number of lanes in some areas.
Therefore we decided to create the graph in a different way.
Assuming that more and more cars are being equipped with sensors that allow for accurate tracking of vehicles, the accumulated trajectory data at the given intersection can be used in order to create the aforementioned graph.
Using this data we can infer,  where possible paths can exist and how probable it is that vehicles follow that path.
This method has the advantage that the result will contain not only the correct number of lanes but also all maneuvers that were observed at some point at this intersection.

\captionsetup[sub]{font=footnotesize}

\begin{figure*}
\centering
\begin{subfigure}[h]{0.32\linewidth}
\includegraphics[width=\linewidth]{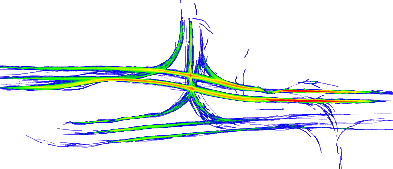}
\caption{Vehicle paths observed at an intersection in Karlsruhe,~Germany.
The color encodes the trajectory density from low (blue) to high (red).}
\label{fig:original picture}
\end{subfigure}
\begin{subfigure}[h]{0.32\linewidth}
\includegraphics[width=\linewidth]{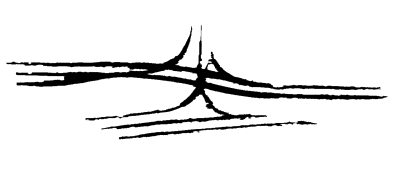}
\caption{Homogenization by binarization of vehicle paths in order to compensate for differing amount of trajectory observations.}
\label{fig:binary picture}
\end{subfigure}
\begin{subfigure}[h]{0.32\linewidth}
\includegraphics[width=\linewidth]{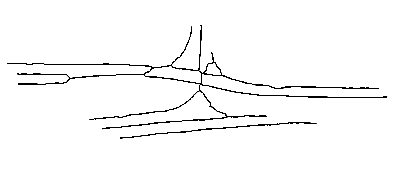}
\caption{Reduction to simple lines through thinning. This depicts the basis for graph extraction.}
\label{fig:picture after thining}
\end{subfigure}

\begin{subfigure}[h]{0.49\linewidth}
\includegraphics[width=\linewidth]{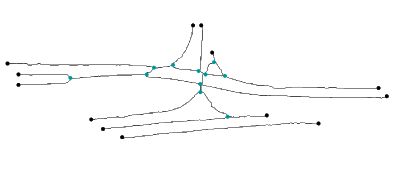}
\caption{Classification of nodes.
Start and end nodes depicted in black, decision and crossover nodes in cyan.}
\label{fig:classification of nodes}
\end{subfigure}
\begin{subfigure}[h]{0.49\linewidth}
\includegraphics[width=\linewidth]{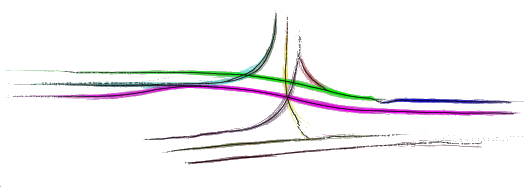}
\caption{Final result containing trajectory prototypes for each velocity cluster which represent different maneuvers.}
\label{fig:trajectories clusters}
\end{subfigure}
\caption{Graph extraction steps. Starting from an aerial trajectory image nodes and edges are extracted by using binarization, thinning and line following algorithms. Afterwards trajectory prototypes are matched to each edge.}
\label{fig:qualitative_results}
\end{figure*}

In order apply this method first we start by transforming all observed car trajectories into the map frame as depicted for a sample intersection in Figure~\ref{fig:original picture}.
The task of extracting typical trajectories is similar to the problem of line detection in a top view representation of these trajectories.
This inspired the idea to use image processing techniques in order to extract the basic graph structure and afterwards using the association of trajectory data to the graph structure to infer all further information.

The process of the topological graph's generation from trajectory data can be summarized in two steps: (a) Extraction of the topology of $G$ using image processing; (b) Classification of each Node $n \in G$ according to possible maneuvers. 
The following sections will explain these steps in detail.

\subsection{Topology extraction}
\label{subsec:graph_generation}
All recorded trajectories in the observed traffic area are first converted to a top-view gray-scale image, in which the brightness of one pixel depends on the number of trajectories, which cross the area represented by this pixel.
Figure~\ref{fig:original picture} depicts this image using a color map to better illustrate brightness differences.
The goal of the first processing step is to eliminate outliers that do not represent the general structure of the observed area.
These outliers are for example lane changes which are generally possible at any point between neighboring lanes but do not change the underlying structure of separate lanes.
For that step, the gray-scale image is processed multiple times with morphological operations (Opening and Closing) with varying parameters.
Since we assume that most irrelevant outlier trajectories were eliminated in the previous step, we assume that the remaining trajectories are all relevant for the graph.
Therefore the next step is the binarization of the image which represents a homogenization of trajectory density as seen in Figure~\ref{fig:binary picture}.
In order to extract possible edges and nodes of the graph the next step is line thinning.
We apply the Zhan-Suen-Thinning algorithm \cite{Zhang:1984} to the binary image and receive a new image in which all lines have a thickness of only one pixel.
This method however might produce artifacts in areas with many diverging trajectories.
These artifacts are very short lines orthogonal to a longer line and can easily be identified and removed by searching for very short branches.
The result of these steps is depicted in Figure~\ref{fig:picture after thining}.
Finally, the nodes and edges are found by searching for black pixels with either only one or more than two black pixels in its direct neighborhood.
Pixels with exactly two black pixels in their neighborhood on the other hand are part of an edge that connects one node with another.
By following all black neighboring pixels of a node to the next node the edges are found.
The set of all nodes and edges found this way generate the structure of $G$. At this point edges are added bidirectionally.

\subsection{Maneuver Classification}
\label{subsec:maneuver_classification}
The determined main structure of topological graph $G$ only contains geometric information about the observed traffic area.
However, the possible movement directions as well as transition probabilities are still missing.
In order to add that information to $G$, map matching is used.
Each recorded trajectory $T$ is matched to an ordered set $E_T = \{e_1, e_2,...e_n\}, e_i \in E, n \in \mathbb{N}_0$ of connected edges in the topological graph.
The matching is done by iterating over all timestamps of each trajectory $T$.
For each timestamp $t \in T$ the current position is transformed into image coordinates and the edge belonging to the closest black pixels is a new candidate to be added at the end of $E_t$.
Each edge $e$ can only be added to $E_T$ if it is not already the last edge in $E_t$ ($e \neq e_n$) and if it is directly connected to a node connected to the last element ($\exists v \in V : (e,v) \in E \land (v,e_n) \in E$) or if $E_T$ is empty ($||E_T||=0$).
Since at this point $G$ is created bidirectionally this always yields two possible nodes $(n_1,n_2),(n_2,n_1)$.
We only add the edge that is aligned with the trajectories movement direction.
The result of this map matching is a sequence that best describes the trajectory in the context of $G$.
After repeating this for all trajectories, every edge that was never part of $E_t$ is removed resulting in a directed graph $G$.
Each node can then be classified into one of four types of nodes: (a) Start nodes only have outgoing edges.
They depict the border of our mapped region. If they occur in $E_T$ they are always the first element in $e_1$.
They are end nodes in neighboring map sections (b) End nodes have only ingoing edges and only occur at the end of $E_T$.
They constitute the outgoing edges of the mapped area and are start nodes in neighboring sections of the map.
(c) crossover nodes have multiple input nodes and multiple output nodes.
However the incoming edge always decides which outgoing edge is chosen.
These nodes are the result of crossing paths in different directions. 
(d) Decision nodes have an input node and multiple output nodes.
The corresponding transition probabilities to different output nodes are between 0 and 1.
The sum of the probabilities to all output nodes is 1.
These are nodes where different maneuvers diverge.
An example classification can be seen in Figure~\ref{fig:classification of nodes}.

%% file: tex/prediction.tex
\section{Prediction}
\label{sec:Prediction}
The result of the previous section of this paper is a graph which describes the possible maneuvers for vehicles in certain areas of the road structure. This chapter describes how to add information to the graph that can be used to predict trajectories and how to perform said prediction. 

\subsection{Accumulate Prediction Information in Data Structure}
\label{subsec:Accumulation}

The information extraction process can again be split into two steps: (1) Calculating transition probabilities based on observed trajectories (2) Selecting prototype trajectories describing typical behavior.

\subsubsection{Transition Probabilities}
\label{subsubsec:transition_probabilities}
From the last step in chapter \ref{sec:Graph Creation}, it can be seen that the transition probabilities at end nodes and crossover nodes are easily determinable.
The corresponding transition probabilities at decision nodes and start nodes with more than one adjacent edge, on the other hand, are still unknown.
In order to determine the transition probabilities in these cases, we assume a relationship between movement velocity and maneuver probability.
The reasoning here is that a change in maneuvers is often accompanied with a change in velocity.
For example vehicles approaching an intersection with high velocity are unlikely to take a turn while vehicles slowing down in front of a fork in the road will more likely take a turn.
In order to get the relationship between velocity and decisions at a decision node, the different speeds of vehicles are first determined in a given distance before reaching the considered decision node based on the trajectory data.
Subsequently, all velocities are clustered based on a processing chain containing the TRACLUS algorithm \cite{Lee:2007} and Agglomerative Hierarchical Clustering \cite{Kumari_performanceevaluation}.
As a first step, all trajectories are clustered with the TRACLUS algorithm based on their velocity.
Because of the different data density of trajectories in the observed traffic area, it is nearly impossible to find parameters that work in all area of the graph equally good.
In order to improve the clustering performance, all clusters detected by the TRACLUS algorithm are checked and if necessary improved by further applying Agglomerative Hierarchical Clustering.

After clustering all trajectories, the transition probability $tp_i^j$ at a decision node can be calculated for each cluster $i$ and output edge $j$ by using the equation $tp_i^j=n_{i,j}/n_i$, where $n_{i,j}$ is the number of vehicles that have a speed in cluster $i$ and drive from the considered decision node to output node $j$.
$n_i$ is the number of vehicles which have a speed in  cluster $i$ and drive cross the considered decision node.
These determined transition probabilities $tp_i^j$ can then be used for trajectory prediction at the corresponding decision node.

\begin{figure}
\centering
\includegraphics[width=\columnwidth]{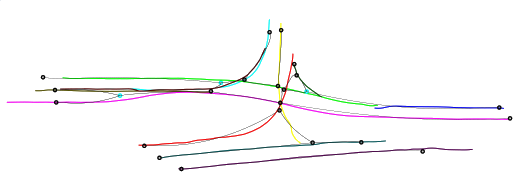}
\caption{Overview over the data used for prediction: Depicted in black are the edges and non-decision nodes, in cyan decision nodes.
Each trajectory color represents a merged prototype trajectory for a given maneuver edge.}
\label{fig:graph and pattern trajectories}
\end{figure}

\subsubsection{Prototype creation}
\label{subsubsec:Prototype_creation}
After all clusters have been determined for each decision node, an attempt is made to extract a prototype trajectory $PT$ from each cluster. 
This extraction is realized using a trajectory density based clustering algorithm proposed by Lee et al. \cite{Lee:2007}. The result of this algorithm is a set of representative trajectories each describing a typical trajectory given a velocity and location within the considered area.
The result is depicted in Figure~\ref{fig:trajectories clusters}).
For each cluster the associated prototype trajectory is stored in the edge of the topological graph $G$ and can be used as a basis for prediction.
In Figure~\ref{fig:graph and pattern trajectories} we illustrate the final resulting data structure with trajectories describing the same maneuver in different velocities merged together for comprehensibility.

\subsection{Prediction Process}
\label{subsec:Prediction_Process}

Our proposed prediction method uses this generated data structure by (1) associating the current movement state to at least one cluster and retrieving the respective prototype trajectories and (2) transforming the prototype trajectories in order to match the vehicle's movement.

\subsubsection{Retrieving trajectory prototypes}
\label{subsubsec:prototype_retrieval}
In order to find prototype trajectories, the current movement state of the vehicle is compared to the edges which represent areas closest to the location of the vehicle. By comparing the direction of edge and movement as well as spatial proximity the best matching edge is found.
If the desired prediction horizon exceeds the length of the respective edge, continuation edges are concatenated based on possible movements through the graph. 
If one of the movements traverses over a decision node this results in several possible edge sequences which can then be processed to several different predictions.
Using these sequences and their corresponding prototype trajectories $\{PT\}$ we can calculate multiple motion predictions with probabilities: 
First we determine the velocity $v_{m}$ of the observed vehicle and compare it to the centers of the velocity clusters for the respective decision node. The result are two neighboring clusters with their respective velocities $v_{slow}$ and $v_{fast}$.
The differences $\delta_{slow}$ and $\delta_{fast}$ between the two neighboring cluster centers and $v_{m}$ are calculated with the equation \eqref{eq:c3 1:a}.
With the difference $\delta$, defined by equation \eqref{eq:c3 1:b} and the corresponding transition probabilities $p_{slow, h}$ and $p_{fast, h}$, the transition probability $p_h$ to the following node $h$ can be calculated using the equation \eqref{eq:c3 1:c}. This results in one or more edge sequences which each represent one prediction and their respective probabilities.
\begin{align}
\label{eq:c3 1:a}
\delta_{\text{slow/fast}} &= \vert v_{\text{m}} - v_{\text{slow/fast}} \vert \\
\label{eq:c3 1:b}
\delta &= \vert v_{\text{fast}} - v_{\text{slow}} \vert\\
\label{eq:c3 1:c}
p_{\text{h}} &= p_{\text{slow}, h} \cdot \frac{\delta_{\text{fast}}}{\delta} + p_{\text{fast}, h} \cdot \frac{\delta_{\text{slow}}}{\delta}
\end{align}

\begin{figure}
	\centering
	\includegraphics[width=\columnwidth]{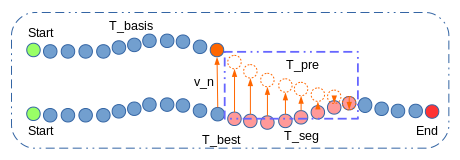}
	\caption{The prototype trajectory is transformed in order to continue the basis trajectory in a sensible way.}
	\label{fig:transformation of trajectories}
\end{figure}

\subsubsection{Transformation of the prototype trajectories}
\label{subsubsec:trajectory_transformation}
In order to calculate the predicted trajectory from a sequence of edges, their respective prototype trajectories and the observed vehicle motion have to be combined to one consistent trajectory.
We are going to explain this process using a simplified example depicted in Figure~\ref{fig:transformation of trajectories}.
The segment $T_\text{seg}$ beginning at the closest point to the observed trajectory is cut out of the respective prototype trajectory $T_{\text{best}}$ with the length depending on the length and distance of each trajectory. This segment $T_{\text{seg}}$ is then transformed to $T_{\text{pre}}$ which connects the observed trajectory with the remainder of the prototype.
In order to obtain a realistic prediction, a spatial transformation of $T_{\text{seg}}$ is required. This transformation can be realized by moving the first point to the end of the observed trajectory and each subsequent point in $T_{\text{seg}}$ into the same direction but by a smaller amount. If we define $v_n$ as the vector between the last point in $T_{best}$ before the beginning of $T_{\text{seg}}$ and the end of the observed Trajectory and the length of $T_{\text{seg}}$ as $n$, then we can use the function $f(i)=1-\frac{i}{n}$ to transform each point $p^{\text{seg}}_i \in T_{\text{seg}}$ to the respective point $p^{\text{pre}}_i \in T_{\text{pre}}$ via equation \ref{eq:c4}. 
\begin{equation}
\label{eq:c4}
p^{\text{pre}}_j = p^{\text{seg}}_j + f(j) \cdot v_n , \quad 1 \leq j \leq n_{seg}
\end{equation}

This transformation is repeated to connect all prototype trajectories in each sequence resulting in a final trajectory each for which we can calculate the probability using equation \ref{eq:c3 1:c}.

%% file: tex/evaluation.tex
\section{Evaluation}
\label{sec:Evaluation}

First, we briefly describe the evaluation data set in Section~\ref{subsec:evaluation_data_set} which provides suitably long trajectories.
Second, we introduce a similarity measure in Section~\ref{subsec:evaluation_similarity_measure} that serves as error measure for our method.
After presenting a baseline method in Section~\ref{subsec:evaluation_baseline_method}, we discuss the results in Section~\ref{subsec:evaluation_discussion}.

\subsection{Data Set}
\label{subsec:evaluation_data_set}

The data set was recorded at a busy intersection in Karlsruhe,~Germany with multiple maneuver options for each lane on two different days and viewpoints to minimize blind spots.

We used a Velodyne HDL64E-S2 \cite{velodyneHDL} range sensor mounted on an experimental vehicle to measure highly accurate surface positions of the environment represented as point sets.
Then, we remove points close to the ground by fitting a RANSAC \cite{fischler1981random} plane model to the data and removing all points below a minimum signed distance.
Afterwards, we perform cluster segmentation based on a generalized Connected Components Labeling on nearest neighbors as presented in \cite{Rusu_ICRA2011_PCL}.
Given the point clusters corresponding to objects we compute their convex hull and assume the object position to be the geometric center of that convex hull.
The tracking is realized by assuming a CYRA motion model and using a simple nearest neighbor association approach as long as each object was visible.
In the next step each proposed object is verified and labeled by hand in order to eliminate objects that are neither cars nor trucks.
Finally, all trajectories are split and/or trimmed so that their segments each are long enough to provide data for different evaluation horizons.

This results in 14531 vehicle trajectories of at least 4 m length and 5766 trajectories being at least 20 m long.

\subsection{Similarity Measure}
\label{subsec:evaluation_similarity_measure}
Choosing an evaluation metric for trajectory prediction while avoiding bias is a challenging task.
In general, a suitable metric should take the predicted trajectorie(s) and compare them with the actual driven trajectory resulting in a value representing either the distance or the similarity to the ground truth.
As detailed in \cite{zheng2015trajectory}, there are several different metrics comparing different aspects of trajectory similarity with each metric providing different results.
Most notably there is a fundamental difference in doing a path or a trajectory comparison.
As proposed in \cite{quehl2017how} it is possible to combine different metrics, each concentrating on different aspects of trajectory similarity, to a weighted sum of several metrics.
In the context of this paper the chosen similarity measure was created based on that proposed method using global orientation difference $m_{\mathrm{GOD}}$, average velocity difference $m_{\mathrm{AVD}}$ and mean euclidean distance for trajectories $m_{\mathrm{MEDT}}$ and paths $m_{\mathrm{MEDP}}$ as basis.
The results showed that $m_{\mathrm{MEDT}}$ and $m_{\mathrm{MEDP}}$ had similar and far larger weights than the other two metrics.
Therefore we chose $m(T_1,T_2) = 0.5 \cdot m_{\mathrm{MEDT}} + 0.5 \cdot m_{\mathrm{MEDP}}$.
This implies that the measure can be seen as the average distance (in meters) from the points closest in time and space. 

\subsection{Baseline Method}
\label{subsec:evaluation_baseline_method}
As a baseline for our evaluation we chose to implement a simple CYRA-based prediction approach.
For this model, we assume constant yaw rate and linear acceleration during prediction.
The resulting trajectory is then compared to the ground truth using the metric defined in Section~\ref{subsec:evaluation_similarity_measure}.
The same procedure was conducted with our graph based prediction approach.
Since our approach also allows to predict more than one trajectory and assign each trajectory an estimated probability we can define this as a third prediction approach for the comparison.
Since this third approach provides several trajectories while the other two only one, the comparison method has to be modified.
For the third approach the evaluation was done by calculating the mathematical expectation based on the estimated probabilities. 

\begin{figure*}
\centering
\includegraphics[width=0.87\textwidth]{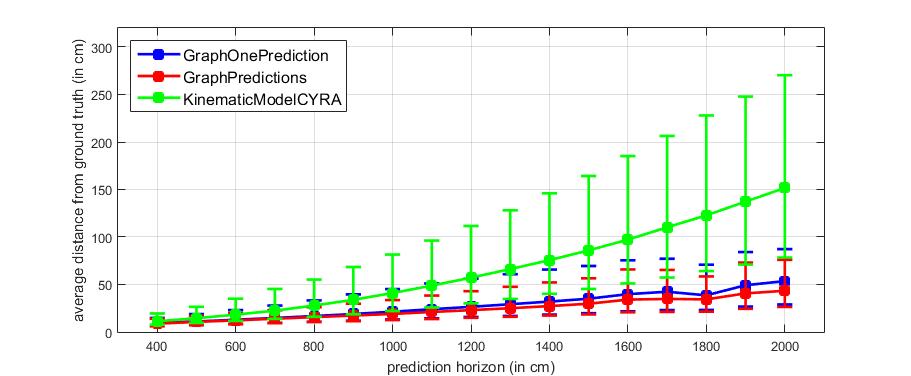}
\caption{Evaluation of our prediction approach (blue and red) compared to prediction with a CYRA motion model (green).
The upper and lower boundaries represent the 75 and 25 percentile, respectively.}
\label{fig:comparison}
\end{figure*}

\subsection{Discussion}
\label{subsec:evaluation_discussion}

The results of our trajectory prediction evaluation are depicted in Fig.~\ref{fig:comparison}.
As previously explained, the evaluation was performed by calculating the predicted position for each timestamp that occurs in the ground truth until a given distance was traveled.
For each point we calculated the mean between the euclidean distance to the closest point on the interpolated ground truth trajectory and the euclidean distance to the point at the same timestamp.
This form of evaluation was chosen in order to mitigate differences in velocity between different trajectories of the data set.
Choosing the prediction horizon based on time difference would instead result in less meaningful data since a second may constitute large spatial differences based on the velocity.

The evaluation shows that both variants of the newly proposed prediction method perform better than a CYRA model based prediction.
While the CYRA assumption seems to hold for small distances, after a few meters it diverges increasingly fast from the ground truth.
The reason for this is that most driving maneuvers at an intersection are completed after a few meters driven causing a change in yaw rate and acceleration.
Our graph based prediction on the other hand incorporates future changes in these parameters.
For example in a situation at the beginning of a curve a CYRA approach predicts that the vehicle will continue driving in a circle or spiral along the current curvature depending on the acceleration.
A graph based approach on the other hand predicts that the vehicle will stop taking the turn after a few meters and continue straight ahead afterwards. 

What is noteworthy is that the average distance to the ground truth of both variants of the graph prediction approach seems to increase only linearly unlike with the CYRA approach.
This implies that in more than $75\%$ of all cases the predicted path was chosen and a small error accumulates linearly with distance traveled.
The fact that the evaluation shows slightly better results for the prediction with more than one result implies that in cases where the most probable estimated trajectory is wrong, a more correct trajectory is still found and assigned a significant probability.

%% file: tex/conclusions.tex
\section{CONCLUSIONS}
\label{sec:Conclusions}
In this paper we introduced a new approach to vehicle trajectory prediction, which utilizes static environment information about drivable maneuvers contained in a map. Our method was evaluated on a set of over 14500 trajectories recorded at an intersection and proved to provide significantly better mid-term prediction results than motion model-based prediction. The second contribution of this paper is a method to automatically create lane-accurate maps from accumulated trajectories and extract a topological graph describing the traffic framework in the considered area.